\def\year{2021}\relax
\documentclass[letterpaper]{article} 
\usepackage{aaai21}  
\usepackage{times}  
\usepackage{helvet} 
\usepackage{courier}  
\usepackage[hyphens]{url}  
\usepackage{graphicx} 
\usepackage{natbib}  
\usepackage{caption} 
\title{Challenges for Using Impact Regularizers to Avoid Negative Side Effects}
\author {
    David Lindner,\textsuperscript{\rm 1}\thanks{The authors contributed equally.}
    Kyle Matoba, \textsuperscript{\rm 2}\footnotemark[1]
    Alexander Meulemans \textsuperscript{\rm 3}\footnotemark[1] \\
}

\affiliations {
    \textsuperscript{\rm 1} Department of Computer Science, ETH Zurich \\
    \textsuperscript{\rm 2} Idiap and EPFL \\
    \textsuperscript{\rm 3} Institute of Neuroinformatics, University of Zurich and ETH Zurich \\
    david.lindner@inf.ethz.ch, kyle.matoba@epfl.ch, ameulema@ethz.ch
}

\usepackage{amsmath}
\usepackage{amssymb}
\usepackage{cleveref}
\usepackage[usenames,dvipsnames]{xcolor}
\usepackage{ifthen}

\newboolean{include-notes}
\setboolean{include-notes}{true}

\newcommand{\TODO}[1]{\ifthenelse{\boolean{include-notes}}{{\color{red} \textbf{TODO: #1}}}{}}

\newcommand{\david}[1]{\ifthenelse{\boolean{include-notes}}{{\color{ForestGreen} \textbf{David}: #1}}{}}

\newcommand{\alex}[1]{\ifthenelse{\boolean{include-notes}}{{\color{Purple} \textbf{Alex}: #1}}{}}

\newcommand{\kyle}[1]{\ifthenelse{\boolean{include-notes}}{{\color{Blue} \textbf{Kyle}: #1}}{}}

\newcommand{\Rtask}{R^{\textnormal{task}}}
\newcommand{\Rframe}{R^{\textnormal{env}}}

\newcommand{\statement}[1]{\textbf{#1}}

\begin{document}

\maketitle
\begin{abstract}
Designing reward functions for reinforcement learning is difficult: besides specifying which behavior is rewarded for a task, the reward also has to discourage undesired outcomes. Misspecified reward functions can lead to unintended negative side effects, and overall unsafe behavior. To overcome this problem, recent work proposed to augment the specified reward function with an impact regularizer that discourages behavior that has a big impact on the environment.
Although initial results with impact regularizers seem promising in mitigating some types of side effects, important challenges remain. In this paper, we examine the main current challenges of impact regularizers and relate them to fundamental design decisions. We discuss in detail which challenges recent approaches address and which remain unsolved. Finally, we explore promising directions to overcome the unsolved challenges in preventing negative side effects with impact regularizers.

\end{abstract}

\section{Introduction}
\label{sec:introduction}
Specifying a reward function in reinforcement learning (RL) that completely aligns with the designer's intent is a difficult task. Besides specifying what is important to solve the task at hand, the designer also needs to specify how the AI system should behave in the environment in general, which is hard to fully cover. For example, RL agents playing video games often learn to achieve a high score without solving the desired task by exploiting the game \citep[e.g.][]{Saunders2017}.
Side effects occur when the behavior of the AI system diverges from the designer's intent because of some considerations that were not anticipated beforehand, such as the possibility to exploit a game. In this work, we focus on side effects that are tied to the reward function, which we define as side effects that would still occur if we had access to an oracle that finds an optimal policy for a given reward function. We explicitly do not consider side effects resulting from the used RL algorithm, which are often discussed using the term \emph{safe exploration} \citep{garcia2015comprehensive}. 

In practice, the designer typically goes through several iterations of reward specification to optimize the agent's performance and minimize side effects. This is often a tedious process and there is no guarantee that the agent will not exhibit side effects when it encounters new situations. In fact, such problems with misspecified reward functions have been observed in various practical applications of RL \citep{krakovnaspecification}.

In most situations, it is useful to decompose the reward $R(s)$ into a task-related component $\Rtask(s)$ and an environment-related component $\Rframe(s)$, where the latter specifies how the agent should behave in the environment, regardless of the task.\footnote{We write the reward function only as a function of states for simplicity, as the state-space can be formally extended to include the last action.} As \citet{Shah2018} observe, $\Rframe$ is related to the frame problem in classical AI \citep{mccarthy1969some}: we not only have to make a prediction about what is supposed to change, but also what is supposed to remain unchanged.
$\Rframe$ is more prone to misspecification, because it needs to specify everything that can happen beyond a task, that can result in undesired outcomes. Because the designer builds an RL agent to solve a specific problem, it is relatively easy to anticipate considerations directly related to solving the task in $\Rtask$.
\citet{Shah2018} point out that environments are generally already optimized for humans, hence, defining $\Rframe$ primarily requires to specify which features of the environment the AI systems should not disturb. Therefore, penalizing large changes in the current state of the world can be thought of as a coarse approximation for $\Rframe$.

Impact regularization (IR) has emerged as a tractable and effective way to approximate $\Rframe$ \citep{armstrong2017low, krakovna2018penalizing, turner2020conservative}. The main idea behind IR is to approximate $\Rframe$ through a measure of ``impact on the environment'', which avoids negative side effects and reduces the burden on the reward designer.

In this paper, we discuss IR of the form
\begin{align}
\label{eqn:regularization_abstract}
R(s_t) = R_{\textnormal{spec}}(s_t) -\lambda \cdot d(s_t, b(s_0, s_{t-1}, t))
\end{align}
where $s_t$ denotes the state at time step $t$, $R_{\textnormal{spec}}$ denotes the reward function specified by the designer,\footnote{$R_{\textnormal{spec}}$ contains the specified parts of both $\Rtask$ and $\Rframe$.} and:

\begin{itemize}
\item the \emph{baseline} $b(s_0, s_{t-1}, t)$ provides a state obtained by following a ``default'' or ``safe'' policy at timestep $t$ and uses either the initial state and the current time $(s_0, t)$ to compute it, or else the current state $s_{t-1}$,
\item $d$ measures the \emph{deviation}
of the realized state from the baseline state, and 
\item  $\lambda \ge 0$ gives a global \emph{scale} at which to trade off the specified reward and the regularization.
\end{itemize}
Composing these three terms gives a general formulation of regularization that encompasses most proposals found in the literature, but permits separate analysis \citep{krakovna2018penalizing}.

We start by giving an overview of the related work on IR (Section \ref{sec:related_work}), before we discuss the three main design decisions for IR.  First, we discuss how to choose a \emph{baseline} (Section \ref{sec:defining_baseline}), emphasizing considerations of environment dynamics and a tendency for agents to offset their actions. 
Second, we discuss how to quantify \emph{deviations} from the baseline (Section \ref{sec:distance_measure}), especially the distinction between negative, neutral, and positive side effects. Third, we discuss how to choose the scale $\lambda$ (Section \ref{sec:magnitude}). Finally, we propose some directions to improve the effectiveness of IR (Section \ref{sec:alternatives}) .

The main contribution of this work is to, discuss in detail the current main challenges of IR, building upon previous work, and to suggest possible ways forward to overcome these challenges.

\section{Related Work}
\label{sec:related_work}

\citet{amodei2016concrete} reviewed negative side effects as one of several problems in AI safety, and discussed using impact regularization (IR) to avoid negative side effects.
Since then, several concrete approaches to IR have been proposed, of which eq. \eqref{eqn:regularization_abstract} gives the underlying structure. \citet{armstrong2017low} proposed to measure the impact of the agent compared to the \textit{inaction baseline}, starting from the initial state $s_0$. The inaction baseline assumes the agent does nothing, which can be formalized by assuming a non-action exists.\footnote{\citet{armstrong2017low} define this baseline as the state the environment would be in when the agent would have never been deployed. This is slightly different from the definition of the inaction baseline we give here and that later work used, as the mere presence of the agent can influence the environment.} \citet{armstrong2017low} emphasized the importance of a semantically meaningful state representation for the environment when measuring distances from the inaction baseline.
While \citet{armstrong2017low} discussed the problem of measuring the impact of an agent abstractly, \citet{krakovna2018penalizing} proposed a concrete deviation measure called \textit{Relative Reachability} (RR). RR measures the average reduction in the number of states reachable from the current state, compared to a baseline state. This captures the intuition that irreversible changes to the environment should be penalized more, but has advantages over directly using irreversibility as a measure of impact (as e.g. in \citet{eysenbach2017leave}), such as allowing to quantify the magnitude of different irreversible changes.

\citet{turner2020conservative} and \citet{krakovna2018penalizing} generalized the concept of RR towards \textit{Attainable Utility Preservation} (AUP) and \textit{Value Difference} (VD) measures respectively, which both share the same structural form for the deviation measure:
\begin{align}\label{eq:value differences}
    d_{\textnormal{VD}}(s_t, s'_t) = \sum_{x=1}^{X} w_x f\big(V_x(s'_t) - V_x(s_t)\big),
\end{align}
where $x$ ranges over some sources of value, $V_x(s_t)$ is the value of state $s_t$ according to $x$, $w_x$ is its weight in the sum and $f$ is a function characterizing the deviation between the values. AUP is a special case of this with $w_x=1/X$ for all $x$ and the absolute value operator as $f$. This formulation captures the same intuition as RR, but allows to measure the impact of the agent in terms of different value functions, instead of just counting states. Concretely, AUP aims to measure the agent's ability to achieve high utility on a range of different goals in the environment, and penalizes any change that reduces this ability.
\citet{turner2020conservative} also introduced the \textit{stepwise inaction baseline} to mitigate offsetting behavior (c.f. Section \ref{sec:offsetting}). This baseline follows an inaction policy starting from the previous state $s_{t-1}$ rather than the starting state $s_0$. Follow-up work scaled AUP towards more complex environments \citep{turner2020avoiding}. 

\citet{krakovna2020avoiding} built upon the VD measure and introduced an auxiliary loss representing how well the agent could solve future tasks in the same environment, given its current state. This can be seen as a deviation measure in e.q. \eqref{eqn:regularization_abstract} that rewards similarity with a baseline instead of penalizing deviation from it. \citet{eysenbach2017leave}'s approach to penalize irreversibility can be seen as a special case of \citet{krakovna2020avoiding}. 

Aside from IR, \citet{rahaman2019learning} proposed to learn an \textit{arrow of time}, representing a directed measure of reachability, using the intuition that irreversible actions tend to leave the environment in a more disorderly state, making it possible to define an arrow of time with methods inspired by thermodynamics. As another alternative to IR, \citet{zhang2018minimax, zhang2020querying} proposed to learn which environmental features an AI system is allowed to change by querying a human overseer. They provided an active querying approach that makes maximally informative queries. \citet{Shah2018} developed a method for learning which parts of the environment a human cares about by assuming that the world is optimized to suit humans. \citet{saisubramanian2020multi} formulated the side effects problem as a multi-objective Markov Decision Process, where they learn a separate reward function penalizing negative side effects and optimize this secondary objective while staying close to the optimal policy of the task objective. \citet{saisubramanian2020avoiding} provide a broad overview of the various existing approaches for mitigating negative side effects, while we zoom in on one class of approaches, IR, and discuss the corresponding challenges in detail.

\section{Choosing a Baseline}
\label{sec:defining_baseline}
Recent work mainly uses two types of baselines in impact regularization (IR): (i) the inaction baseline $b(s_0, s_t, t) = T(s_t | s_0, \pi_{\mathrm{inaction}})$ and (ii) the stepwise inaction baseline $b(s_0, s_t, t) = T(s_t | s_{t-1}, \pi_{\mathrm{inaction}})$, where $T$ is the distribution over states $s_t$ when starting at state $s_0$ or $s_{t-1}$ respectively and following the inaction policy $\pi_{\mathrm{inaction}}$  that always takes an action $a_{\mathrm{nop}}$ that does nothing.

Unfortunately, the inaction baseline can lead to undesirable offsetting behavior, where the agent tries to undo the outcomes of their task after collecting the reward, moving back closer to the initial baseline \citep{turner2020conservative}. The stepwise inaction baseline removes the offsetting incentive of the agent by branching off from the previous state instead of the starting state \citep{turner2020conservative}. However, \citet{krakovna2020avoiding} argued that offsetting behavior is desirable in many cases. In section \ref{sec:offsetting} we contribute to this discussion by breaking down in detail when offsetting behavior is desirable or undesirable, whereas in section \ref{sec:environment_dynamics}, we argue that the inaction baseline and step-wise inaction baseline can lead to \textit{inaction incentives} in nonlinear dynamical environments. We start, however, with the fundamental observation that the inaction baseline and stepwise inaction baseline do not always represent safe policies in section \ref{sec:inact_baseline}.

\subsection{Inaction Baselines are not Always Safe} \label{sec:inact_baseline}
The baseline used in IR should represent a safe policy where the AI system does not harm its environment or itself. In many cases, taking no actions would be a safe policy for the agent, e.g. for a cleaning robot. However, if the AI system is responsible for a task requiring continuous control, inaction of the AI system can be disastrous. For example, if the agent is responsible for driving a car on a highway, doing nothing likely results in a crash. This is particularly problematic for the stepwise inaction baseline, which follows an inaction policy starting from the previous state. The inaction policy starting from the initial state can also be unsafe, for example, if an agent takes over the control of the car from a human, and therefore the initial state $s_0$ already has the car driving.

For this reason, designing a safe baseline for a task or environment that requires continuous control is a hard problem. One possible approach is to design a policy that is known to be safe based on expert knowledge. However, this can be a time-consuming process, and is not always feasible. Designing safe baselines for tasks and environments that require continuous control is an open problem that has to be solved before IR can be used in these applications.

\subsection{Offsetting}\label{sec:offsetting}
An agent engages in offsetting behavior when it tries to undo the outcomes of previous actions, i.e. when it ``covers up its tracks''. Offsetting behavior can be desirable or undesirable, depending on which outcomes the agent counteracts. 

\textbf{Undesirable offsetting.}
Using IRs with an inaction baseline starting from the initial state can lead to undesirable offsetting behavior where the agent counteracts the outcomes of its task \citep{krakovna2018penalizing, turner2020conservative}. 
For example, \citet{krakovna2018penalizing} consider a vase on a conveyor belt. The agent is rewarded for taking the vase off the belt, hence preventing that it will fall off the belt. The desired behavior is to take the vase and stay put. The offsetting behavior is to take the vase off the belt, collect the reward, and afterwards put the vase back on the conveyor belt to reduce deviation from the baseline.
To understand this offsetting behavior recall the decomposition of the true reward into a task-related and an environment-related component from section \ref{sec:introduction}. A designer usually specifies a task reward $\Rtask_{\textnormal{spec}}$ that rewards states signaling task completion (e.g. taking the vase off the belt). However, each task has consequences to the environment, which often are the reason why the task should be completed in the first place (e.g. the vase being not broken). In all but simple tasks, assigning a reward to every task consequence is impossible, and so by omission, they have a zero reward. When IR penalizes consequences of completing the task, because they differ from the baseline, this results in undesirable offsetting behavior. The stepwise inaction baseline \citep{turner2020avoiding} successfully removes all offsetting incentives. However, in other situations offsetting might be desired.

\textbf{Desirable Offsetting.}
In many cases, offsetting behavior is desired, because it can prevent unnecessary side effects. \citet{krakovna2020avoiding} provide an example of an agent which is asked to go shopping, and needs to open the front door of the house to go to the shop. If the agent leaves the door open, wind from outside can knock over a vase inside, which the agent can prevent by closing the door after leaving the house. When using the stepwise inaction baseline (with rollouts, c.f. Section \ref{sec:rollout_policies}), the agent gets penalized once when opening the door for knocking over the vase in the future, independent of whether it closes the door afterwards (and thus prevents the vase from breaking) or not. Hence, for this example, the offsetting behavior (closing the door) is desirable. The reasoning behind this example can be generalized to all cases where the offsetting behavior concerns states that are instrumental towards achieving the task (e.g. opening the door) and not a consequence of completing the task (e.g. the vase being not broken).

\textbf{A Crucial Need for a New Baseline.} The recently proposed baselines either remove offsetting incentives altogether or allow for both undesirable and desirable offsetting to occur, which are both unsatisfactory solutions. \citet{krakovna2020avoiding} proposed resolving this issue by allowing all offsetting (e.g. by using the inaction baseline) and rewarding all states where the task is completed in the specified reward function. However, we attribute three important downsides to this approach. First, states that occur after task completion can still have negative side effects. If the reward associated with these states is high enough to prevent offsetting, it might also be high enough to encourage the agent to pursue these states and ignore their negative side effects. Second, not all tasks have a distinct goal state that indicates the completion of a task, but rather accumulate task-related rewards at various time steps during an episode. Third, this approach creates a new incentive for the agent to prevent shut-down, as it continues to get rewards after the task is completed \citep{hadfield2017off}.

We conclude that offsetting is still an unsolved problem, highlighting the need for a new baseline, to prevent undesirable offsetting behavior, but allow for desirable offsetting.

\subsection{Environment Dynamics and Inaction Incentives} \label{sec:environment_dynamics}
In dynamic environments that are highly sensitive to the agent's actions, the agent will be susceptible to \textit{inaction incentives}. Either the agent does not act at all (for all but small magnitudes of $\lambda$) or it will be insufficiently regularized and possibly result in undesired side effects (for small $\lambda$). 

\textbf{Sensitivity to Typical Actions.} Many real-world environments exhibit chaotic behavior, in which the state of the environment is highly sensitive to small perturbations. In such environments, the environment state where the agent has performed an action will be fundamentally different from the environment state for the inaction baseline \citep{armstrong2017low}. Furthermore, for the step-wise inaction baseline, the same argument holds for the non-action compared to the planned action of the agent. Hence, when using these baselines for IR, all actions of the agent will be strongly regularized, creating the inaction incentive. When $\lambda$ is lowered to allow the agent to take actions, the agent can cause negative side effects when the IR cannot differentiate between negative side effects and chaotic changes in the environment. Here, it is useful to distinguish between \emph{typical} and \emph{atypical} actions. We say (informally) that an action is \emph{typical} if it is commonly used for solving a wide variety of tasks (e.g. moving). When the environment is highly sensitive to typical actions, IRs with the current baselines will prevent the agent from engaging in normal operations. However, it is not always a problem if the environment is highly sensitive to atypical actions of the agent (e.g. discharging onboard weaponry), as preventing atypical actions impedes less with the normal operation of the agent.

\textbf{Capability of the Agent.} The inaction incentive will become more apparent for agents that are highly capable of predicting the detailed consequences of their actions, for example by using a powerful physics engine. As the ability to predict the consequences of an action is fundamental to minimizing side effects, limiting the prediction capabilities of an agent to prevent the inaction incentive is not desired. Rather, for agents that can very accurately predict the implications of their actions, it is necessary to have an accompanying intelligent impact regularizer.

\textbf{State Features.}
\citet{armstrong2017low} point out that for IR one should not represent states with overly fine-grained features, as presenting an agent with too much information exposes them to basing decisions on irrelevancies.
For example, it would be counterproductive for an agent attempting to forecast demand in an online sales situation to model each potential customer separately, when broader aggregates would suffice. However, there remain two issues with this approach to mitigate the inaction incentive. First, the intrinsic dynamics of the environment remain unchanged, so it is still highly sensitive to small perturbations, of which the results can be visible in the coarser features (e.g. the specific weather conditions). Second, for advanced AI systems, it might be beneficial to change their feature representation to become more capable of predicting the consequences of their actions. In this case, one would have no control over the granularity of the features.

\textbf{Deviation Measures.} At the core of the inaction problem is that some negative side effects are worse than others. Usually, it does not matter if the agent changes the weather conditions by moving around, however, it would matter if the agent causes a serious negative side effect, for example a hurricane. While both outcomes can be a result of complex and chaotic dynamics of the environment, we care less about the former and more about the latter. Differentiating between negative, neutral and positive side effects is a task of the deviation measure used in the IR, which is discussed in the next section.

\section{Choosing a Deviation Measure}
\label{sec:distance_measure}
A baseline defines a ``safe'' counterfactual to the agent's actions. The deviation measure determines how much a deviation from this baseline by the agent should be penalized or rewarded. Currently, the main approaches to a deviation measure are the relative reachability (RR) measure \citep{krakovna2018penalizing}, the attainable utility preservation (AUP) measure \citep{turner2020conservative} and the future task (FT) reward \citep{krakovna2020avoiding}. AUP and FT still require a specification of which tasks the agent might want to achieve in future. In this section, we argue that the current deviation measures still require to specify a notion of \emph{value} of the impact to avoid unsatisfactory performance of the agent and that new rollout policies should be designed that allow for a proper incorporation of delayed effects into the deviation measure.

\subsection{Which Side Effects are Negative?}
The goal of IRs is to approximate $\Rframe$ for all states in a tractable manner. It does this by penalizing impact on the environment, built upon the assumption that the environment is already optimized for human preferences \citep{Shah2018}. 
The IR aims to penalize impact proportionally to the magnitude of this impact which corresponds with the magnitude of the side effect \citep{krakovna2018penalizing, turner2020conservative}. However, not all impact is negative, but it can also be neutral or even positive. $\Rframe$ does not only consider the magnitude the impact on the environment, but also to which degree this impact is negative, neutral or positive. Neglecting the associated value of impact can lead to suboptimal agent behavior, as highlighted in the example below.

\statement{Example: The Chemical Production Plant.}
\label{sec:chemical_production}
Consider an AI system controlling a plant producing a chemical product for which various unknown reactions exist, each producing a different combination of waste products. The task of the AI system is to optimize the production rate of the plant, i.e. it gets a reward proportional to the production rate. To minimize the impact of the plant on the environment, the reward function of the agent is augmented with an impact regularizer, which penalizes the mass of waste products released in the environment, compared to an inaction baseline (where the plant is not operational). Some waste products are harmless (e.g. $O_2$), whereas others can be toxic. When the deviation measure of the impact regularizer does not differentiate between negative, neutral or positive impact, the AI system is incentivized to use a reaction mechanism that maximizes production while minimizing waste. However, this reaction might output mostly toxic waste product, whereas another reaction outputs only harmless waste products and hence has no negative side effects. Tuning the regularizer magnitude $\lambda$ does not provide a satisfactory solution in this case, as either the plant is not operational (for high lambda), or the plant is at risk of releasing toxic waste products in the environment.

\statement{Positive Side Effects.}
The distinction between positive, neutral and negative impact is not only needed to allow for a satisfactory performance of the agent in many environments, it is also desirable for encouraging unanticipated positive side effects. Expanding upon the example in \ref{sec:chemical_production}: if the agent discovered a way to costlessly sequester carbon dioxide alongside its other tasks it should do so, whilst an IR would encourage the robot to not interfere.
While very positive unexpected outcomes might be unlikely, this possibility should not be neglected in the analysis of impact regularizers. 

\statement{Value Differences.}
To distinguish between positive, neutral and negative side effects, we need an approximation of $\Rframe$ that goes beyond measuring impact as a sole source of information. Attainable utility preservation \citep{turner2020conservative} allows for differentiating between positive and negative impact by defining the deviation measure as a sum of differences in value between a baseline and the agent's state-action pair for various value functions. Hence, it is possible to reflect how much the designer's values different kinds of side effects in these value functions.
However, the challenge remains to design value functions that approximate $\Rframe$ to a sufficient degree on the complete state space, which is again prone to reward misspecification. So although the value difference framework allows for specifying values for side effects, \emph{how} to specify this notion of value is still an open problem. 
\subsection{Rollout Policies}\label{sec:rollout_policies}
Often, the actions of an agent cause delayed effects, i.e. effects that are not visible immediately after taking the action. The stepwise inaction baseline \citep{turner2020conservative} ignores all actions that took place before $t-1$, hence, to correctly penalize delayed effects, the deviation measure needs to incorporate future effects. This can be done by collecting rollouts of future trajectories using a simulator or model of the environment. These rollouts depend on which \emph{rollout policy} is followed by the agent in the simulation. For the baseline states, the inaction policy is the logical choice. For the rollout of the future effects of the agent's action, it is less clear which rollout policy should be used. \citet{turner2020conservative} use the inaction policy in this case. Hence, this IR considers a rollout where the agent takes its current action, after which it cannot do any further actions. This approach has significant downsides, because IR does not allow the agent to do a series of actions when determining the impact penalty (e.g. the agent can take an action to jump, but cannot plan for its landing accordingly in the rollout). Therefore, we argue that future work should develop rollout policies different from the inaction policy, such as the current policy of the agent.

\section{Choosing the Magnitude of the Regularizer}
\label{sec:magnitude}

To combine the IR with a specified reward function, the designer has to choose the magnitude of the regularizer $\lambda$. \citet{turner2020conservative} say that ``loosely speaking, $\lambda$ can be interpreted as expressing the designer's beliefs about the extent to which $R$ [the specified reward] might be misspecified''.

It is crucial to choose the correct $\lambda$. If $\lambda$ is too small, the regularizer may not reduce the risk of undesirable side effects effectively. If $\lambda$ is too big, the regularizer will overly restrict necessary effects of the agent on the environment, and the agent will be less effective at achieving its goal. Note, that while the regularizers proposed by \citet{krakovna2018penalizing} and \citet{turner2020conservative} already measure utility, in general $\lambda$ must also handle a unit-conversion of the regularizer to make it comparable with the reward function.

Some intuition for choosing $\lambda$ comes from a Bayesian perspective, where the regularizer encodes prior knowledge and $\lambda$ controls how far from the prior the posterior should have moved. Another distinct view on setting $\lambda$ comes from the dual optimization problem, where it represents the Lagrange multiplier on an implied set of constraints: $\lambda$ is the magnitude of the regularizer for which the solution to the penalized optimization problem coincides with a constrained optimization problem. Hence, the designer can use $\lambda$ to communicate constraints to the AI system, which is a natural way to phrase some common safety problems \citep{ray2019benchmarking}.

\citet{armstrong2017low} discuss the problem of tuning $\lambda$ and note that contrary to intuition the region of useful $\lambda$'s can be very small and hard to find safely.
In practice $\lambda$ is often tuned until the desired behavior is achieved, e.g., by starting with a high $\lambda$ and reducing it until the agent achieves the desired behavior. This approach is in general insufficient to find the correct trade-off. For a fixed step-size in decreasing $\lambda$, the tuning might always jump from a $\lambda$ that leads to inaction, to a $\lambda$ that yields unsafe behavior. The same holds for other common procedures to tune hyperparameters.

\section{Ways Forward}
\label{sec:alternatives}
In this section, we put forward promising future research directions to overcome the challenges discussed in the previous sections.

\subsection{A Causal Framing of Offsetting} \label{sec:causal_offsetting}
In Section \ref{sec:offsetting}, we highlighted that some offsetting behavior is desired and some undesired. To design an IR that allows for desired offsetting but prevents undesired offsetting, one firsts needs to have a mechanism that can predict and differentiate between these two types of offsetting. Undesired offsetting concerns the environment states that are a consequence of the task. The difficulty lies in determining which states are a causal consequence of the task being completed and differentiate them from states that could have occurred regardless of the task. 

\statement{Goal-based Tasks.} When the task consists of reaching a certain goal state, the consequences of performing a task can be formalized in a causal framework \citep{pearl2009causality}. When a causal graph of the environment-agent-interaction is available, the states that are a consequence of the task can be obtained from the graph as the causal children nodes of the goal state. Hence, a baseline that allows for desired offsetting behavior but prevents undesired offsetting behavior prevents the agent from interfering with the children nodes of the goal states, while allowing for offsetting on other states.

\statement{General Tasks.} Not all tasks have a distinct goal state which indicates the completion of a task, but accumulate instead task-related rewards at various time steps during an episode. Extending this argument to general tasks remains an open issue, for which causal influence diagrams \citep{everitt2019understanding} can provide a mathematical framework. 

\subsection{Probabilities Instead of Counterfactuals as Baseline}
\citet{armstrong2017low} made the interesting argument that probabilities are better suited than counterfactuals for measuring the impact of actions. Current implementations of IRs use a counterfactual as baseline (e.g. the inaction baseline or stepwise inaction baseline). Because this baseline is one specific trajectory, it will differ considerably from the actual trajectory of the agent in environments that exhibit chaotic dynamics. However, chaotic environments will also be highly sensitive to perturbations that do not originate from the agent's actions. One possible way forward towards a more robust measure of the agent's impact on the environment is hence to compare probabilities that marginalize over all external perturbations instead of comparing specific trajectories. Define $p(s_t|A)$ as the probability of having state $s_t$, given the trajectory of actions $A$ the agent took and $p(s_t|B)$ as the probability of $s_t$ given actions prescribed by the baseline. All influences of perturbations that did not arise from the agent are marginalized out in these probabilities. Hence, a divergence measure between these two probabilities can give a more robust measure of potential impact of the agent, without being susceptible to non-necessary inaction incentives.
To our best knowledge, this idea has not yet been implemented as a concrete IR method and would hence be a promising direction for future research.

\subsection{Improved Human-Computer interaction} \label{sec:human_computer}
Side effects occur if there is a difference between the outcome an AI system achieves and the intent of its (human) designer. Thus improving how well the designer can communicate their intent to the AI system is an important aspect of eliminating side effects \cite{leike2018scalable}. This emphasis on the human component of learning to avoid negative side effects connects it closely to the problem of \emph{scalable oversight} proposed by \citet{amodei2016concrete}.

\statement{Improved Tools for Reward Designers.} Commonly, a designer will aim to iteratively improve the AI system and its reward function. Similarly, when choosing an impact regularizer, a designer will iterate on the choice of baseline, deviation measure, and regularization strength and test them in a sequence of environments that increasingly resemble the production environment. At each iteration, the designer identifies weaknesses and corrects them, such that the criterion being optimized becomes increasingly true to the designer's intent. For example, an AI with the goal to trade financial assets may be run against historical data (``backtested'') in order to understand how it might have reacted in the past, and presented with deliberately extreme inputs (``stress-tested'') in order to understand likely behavior in ``out of sample'' situations. To design a reward function and a regularizer, it is crucial for the designer to be able to understand how the system would react in novel situations and how to fix it in case it exhibits undesired behavior. Further research aiming to increase the designer's ability to understand how a system will react, will substantially help the designer to communicate their intent more effectively. Recent work in this direction concerning \emph{interpretability} \citep{gilpin2018explaining}, \emph{verification} \citep[e.g.][]{huang2017safety} of machine learning models is particularly promising.

\statement{Actively Learning from Humans.}
Considering the problem from the perspective of the AI system, the goal is to improve its ability to understand the designer's intent, especially in novel, unanticipated, scenarios. Instead of the designer \emph{telling} the system their intent, this problem can be addressed by the system \emph{asking} the designer about their intent. To decide what to ask the designer, the system may be able to determine which states it is highly uncertain about, even if it is not able to accurately ascribe values to some of them. Recent work shows that such an approach can be effectively used to learn from the human about a task at hand \citep{christiano2017deep}, but it may also be used to learn something about the constraints of the environment and which side effects are desired or undesired \citep{zhang2018minimax}. Active learning could also provide a different perspective on impact regularizers: instead of directly penalizing impact on the environment, a high value of the regularization term could be understood as indicating that the designer should give feedback. In particular, this approach could help to resolve situations in which a positive task reward conflicts with the regularization term.

\section{Conclusion}

Avoiding negative side effects in systems that have the capacity to cause harm is necessary to fully realize the promise of artificial intelligence. In this paper, we discussed a popular approach to reduce negative side effects in RL: impact regularization (IR). We discussed the practical difficulty of choosing each of the three components: a baseline, a deviation measure and a regularization strength. Furthermore, we pointed to fundamental problems that are currently not addressed by state-of-the-art methods, and presented several new future research directions to address these. While our discussion showed that current approaches still leave significant opportunities for future work, IRs are a promising idea for building the next generation of safe AI systems, and we hope that our discussion is valuable for researchers trying to build new IRs.

\section*{Acknowledgments}
We thank Andreas Krause, Fran\c{c}ois Fleuret and Benjamin Grewe for their valuable comments and suggestions. Kyle Matoba was supported by the Swiss National Science Foundation under grant number FNS-188758 ``CORTI''.

\newcommand{\ICML}{Proceedings of International Conference on Machine Learning (ICML)}
\newcommand{\RSS}{Proceedings of Robotics: Science and Systems (RSS)}
\newcommand{\NeurIPS}{Advances in Neural Information Processing Systems}
\newcommand{\IJCAI}{Proceedings of International Joint Conferences on Artificial Intelligence (IJCAI)}
\newcommand{\ICLR}{International Conference on Learning Representations (ICLR)}
\newcommand{\CoRL}{Conference on Robot Learning (CoRL)}
\newcommand{\AAAI}{Proceedings of the AAAI Conference on Artificial Intelligence}
\bibliography{references}

\clearpage
\appendix

\end{document}